\documentclass[letterpaper]{article} 
\usepackage{aaai}
\pdfoutput=1
\usepackage{times}
\usepackage{helvet} 
\usepackage{courier}
\usepackage{graphicx}
\usepackage[lofdepth,lotdepth]{subfig}

\graphicspath{{img//}}
\title{CAPIR: Collaborative Action Planning with Intention Recognition}

\author{Truong-Huy Dinh Nguyen \and David Hsu \and Wee-Sun Lee \and Tze-Yun Leong \\ 
Department of Computer Science, National University of Singapore, Singapore 117417, Singapore
\\ \{trhuy,dyhsu,leews,leongty\}@comp.nus.edu.sg
\AND Leslie Pack Kaelbling \and Tomas Lozano-Perez \\ 
MIT Computer Science and Artiﬁcial Intelligence Laboratory, Cambridge, MA 02139, USA
\\ \{lpk,tlp\}@csail.mit.edu
\AND Andrew Haydn Grant  \\
Singapore-MIT GAMBIT Game Lab, Cambridge, MA 02139, USA
\\ haydn@mit.edu}

\begin{document}
\copyrightyear{2011}
\maketitle

\begin{abstract}
We apply decision theoretic techniques to construct non-player characters that are able to assist a human player in collaborative games. The method is based on solving Markov decision processes, which can be difficult when the game state is described by many variables. To scale to more complex games, the method allows decomposition of a game task into subtasks, each of which can be modelled by a Markov decision process. Intention recognition is used to infer the subtask that the human is currently performing, allowing the helper to assist the human in performing the correct task. Experiments show that the method can be effective, giving near-human level performance in helping a human in a collaborative game.  
\end{abstract} 

\section{Introduction} 
\label{sec:intro}

Traditionally, the behaviour of Non-Player Characters (NPCs) in games is hand-crafted by programmers using techniques such as Hierarchical Finite State Machines (HFSMs) and Behavior Trees \cite{champ2007}. These techniques sometimes suffer from poor behavior in scenarios that have not been anticipated by the programmer during game construction. In contrast, techniques such as Hierarchical Task Networks (HTNs) or Goal-Oriented Action Planner (GOAP)~\cite{orkin2004} specify goals for the NPCs and use planning techniques to search for appropriate actions, alleviating some of the difficulties of having to anticipate all possible scenarios.

In this paper, we study the problem of creating NPCs that are able to help players play collaborative games. The main difficulties in creating NPC helpers are to understand the intention of the human player and to work out how to assist the player. Given the successes of planning approaches to simplifying game creation, we examine the application of planning techniques to the collaborative NPC creation problem.  In particular, we extend a decision-theoretic framework for assistance used in \cite{fern2010computational} to make it appropriate for game construction.

The framework in \cite{fern2010computational} assumes that the computer agent needs to help the human complete an unknown task, where the task is modeled as a Markov decision process (MDP)~\cite{Bel}. The use of MDPs provide several advantages such as the ability to model noisy human actions and stochastic environments. Furthermore, it allows the human player to be modelled as a noisy utility maximization agent where the player is more likely to select actions that has high utility for successfully completing the task. Finally, the formulation allows the use of Bayesian inference for intention recognition and expected utility maximization in order to select the best assistive action. 

Unfortunately, direct application of this approach to games is limited by the size of the MDP model, which grows exponentially with the number of characters in a game. To deal with this problem, we extend the framework to allow decomposition of a task into subtasks, where each subtask has manageable complexity.  Instead of inferring the task that the human is trying to achieve, we use intention recognition to infer the current subtask and track the player's intention as the intended subtask changes through time.

For games that can be decomposed into sufficiently small subtasks, the resulting system can be run very efficiently in real time. We perform experiments on a simple collaborative game and demonstrate that the technique gives competitive performance compared to an expert human playing as the assistant.

\section{Scalable Decision Theoretic Framework}
\begin{figure}[h]
\centering
\includegraphics[height=120pt]{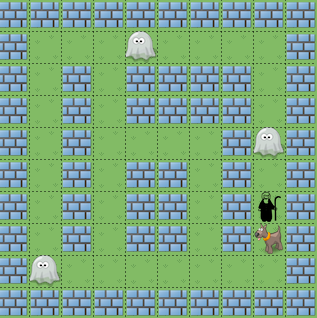} 
\caption{A typical level of Collaborative Ghostbuster. The protagonists, Shepherd and Dog in the bottom right corner, need to kill all three ghosts to pass the level.}
\label{fig:sample_gb1}
\end{figure}

We will use the following simple game as a running example, as well as for the experiments on the effectiveness of the framework. In this game, called \textit{Collaborative Ghostbuster}, the assistant (illustrated as a dog) has to help the human kill several ghosts in a maze-like environment. A ghost will run away from the human or assistant when they are within its vision limit, otherwise it will move randomly. Since ghosts can only be shot by the human player, the dog's role is strictly to round them up. The game is shown in Figure~\ref{fig:sample_gb1}. Note that collaboration is often truly required in this game - without surrounding a ghost with both players in order to cut off its escape paths, ghost capturing can be quite difficult.

\subsection{Markov Decision Processes}
We first describe a Markov decision process and illustrate it with a Collaborative Ghostbuster game that has a single ghost.

A Markov decision process is described by a tuple $ (S, A, T, R) $  in which

\begin{itemize}
\item $ S $ is a finite set of game states. \textit{In single ghost Collaborative Ghostbuster, the state consists of the positions of the human player, the assistant and the ghost.} 
\item $ A $ is a finite set of actions available to the players; each action $ a \in A $  could be a compound action of both players. \textit{If each of the human player and the assistant has 4 moves (north, south, east and west), $A$ would consist of the 16 possible combination of both players' moves.}
\item $ T_{a}(s, s') = P(s_{t+1}=s' \vert s_{t}=s, a_{t}=a) $ is the probability that action $ a $ in state $ s $ at time $ t $ will lead to state $s'$ at time $ t+1 $. \textit{The human and assistant move deterministically in Collaborative Ghostbuster but the ghost may move to a random position if there are no agents near it.}
\item $ R_{a}(s,s')$ is the immediate reward received after the state transition from $s$ to $s'$ triggered by action $a$. \textit{In Collaborative Ghostbuster, a non-zero reward is given only if the ghost is killed in that move. }
\end{itemize}

The aim of solving an MDP is to obtain a policy $\pi$ that maximizes the expected cumulative reward $\sum_{t=0}^{\infty}\gamma^{t}R_{\pi(s_{t})}(s_{t}, s_{t+1})$ where $0 < \gamma < 1$ is the discount factor.

\subsubsection*{Value Iteration.} 
An MDP can be effectively solved using a simple algorithm proposed by Bellman in 1957~\cite{Bel}. The algorithm maintains a value function $V(s)$, where $s$ is a state, and iteratively updates the value function using the equation
\[ V_{t+1}(s)=\max_{a}\left(\sum_{s'}T_{a}(s,s')(R_{a}(s,s')+\gamma V_{t}(s'))\right). \]
This algorithm is guaranteed to converge to the optimal value function $V^*(s)$, which gives the expected cumulative reward of running the optimal policy from state $s$.

The optimal value function $V^{*}$ can be used to construct the optimal actions by taking action $a^{*}$ in state $s$ such that $a^{*}=\mbox{argmax}_{a} \left\lbrace \sum_{s'} T_{a}(s,s')V^{*}(s') \right\rbrace$. The optimal $Q$-function is constructed from $V^{*}$ as follows:
\[ Q^{*}(s,a) = \sum_{s'} T_{a}(s,s')(R_{a}(s,s') + \gamma V^{*}(s')).\]
The function $Q^{*}(s,a)$ denotes the maximum expected long-term reward of an action $a$ when executed in state $s$ instead of just telling how valuable a state is, as does $V^{*}$. 

\subsubsection*{Intractability.} One key issue that hinders MDPs from being widely used in real-life planning tasks is the large state space size (usually exponential in the number of state variables) that is often required to model realistic problems.

Typically in game domains, a state needs to capture all essential aspects of the current configuration and may contain a large number of state variables. For instance, in a Collaborative Ghostbuster game with a maze of size $m$ (number of valid positions) consisting of a player, an assistant and $n$ ghosts, the set of states is of size $O(m^{n+2})$, which grows exponentially with the number of ghosts.

\subsection{Subtasks}
To handle the exponentially large state space, we decompose a task into smaller subtasks and use intention recognition to track the current subtask that the player is trying to complete.

\begin{figure}[h]
\centering
\includegraphics[scale=0.5]{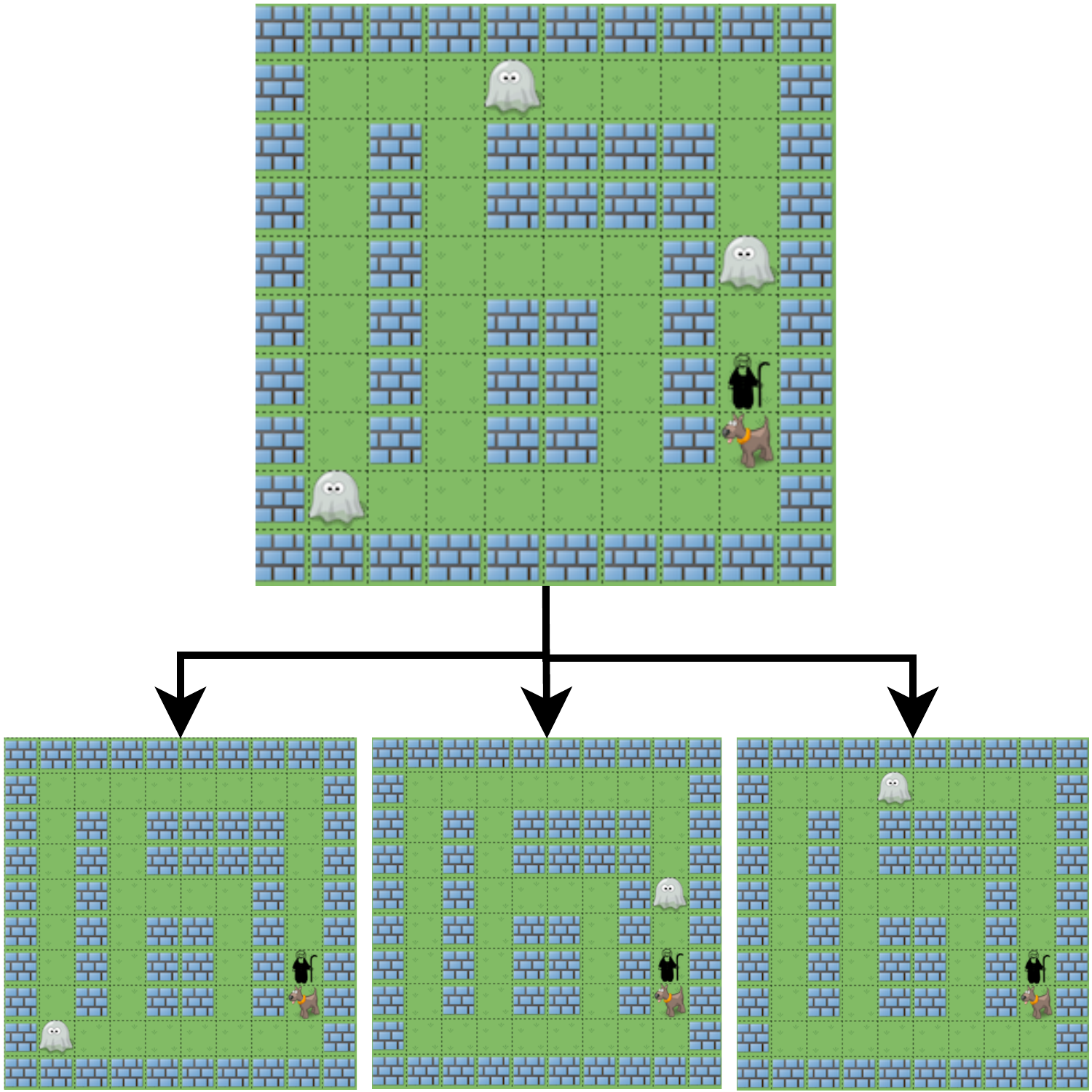} 
\caption{Task decomposition in Collaborative Ghostbuster.}
\label{fig:subworld_decomp1}
\end{figure}

In Collaborative Ghostbuster, each subtask is the task of catching a single ghost, as shown in Figure~\ref{fig:subworld_decomp1}. The MDP for a subtask consists of only two players and a ghost and hence has manageable complexity. 

\subsubsection{Human Model of Action Selection}

In order to assist effectively, the AI agent must know how the human is going to act. Without this knowledge, it is almost impossible for the AI to provide any help. We assume that the human is mostly rational and use the $Q$-function to model the likely human actions.

Specifically, we assume
\begin{equation}
\label{update_actsel}
P(a_{human} \vert w_{i}, s_i) = \alpha . e^{max_{a_{AI}} Q_{i}^{*}(s_i, a_{human}, a_{AI})}
\end{equation}
where $\alpha$ is the normalizing constant, $w_i$ represents subtask $i$ and $s_i$ is the state in subtask $i$. Note that we assume that the human player knows the best response from the AI sidekick and plays his part in choosing the action that matches the most valued action pair. However, the human action selection can be noisy, as modelled by Equation~(\ref{update_actsel}).

\subsection{Intention Recognition and Tracking}

\begin{figure}[h]
\centering
\includegraphics[height=108pt]{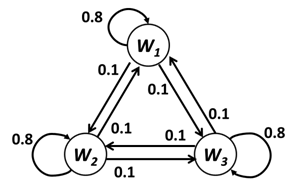} 
\caption{A probabilistic state machine, modeling the transitions between subtasks.}
\label{fig:hum_model}
\end{figure}

We use a probabilistic state machine to model the subtasks for intention recognition and tracking. At each time instance, the player is likely to continue on the subtask that he or she is currently pursuing. However, there is a small probability that the player may decide to switch subtasks. This is illustrated in Figure~\ref{fig:hum_model}, where we model a human player who tends to stick to his chosen sub-goal, choosing to solve the current subtask 80\% of the times and switching to other sub-tasks 20\% of the times. The transition probability distributions of the nodes need not be homogeneous, as the human player could be more interested in solving some specific subtask right after another subtask. For example, if the ghosts need to be captured in a particular order, this constraint can be encoded in the state machine. The model also allows the human to switch back and forth from one subtask to another during the course of the game, modelling change of mind.

\subsubsection{Belief Representation and Update}

The belief at time $t$, denoted $B_{t}(w_{i} \vert \theta_{t})$, where $\theta_{t}$ is the game history, is the conditional probability of that the human is performing subtask $i$. The belief update operator takes $B_{t-1}(w_{i} \vert \theta_{t-1})$ as input and carries out two updating steps. 

First, we obtain the next subtask belief distribution, taking into account the probabilistic state machine model for subtask transition $T(w_{k} \rightarrow w_{i})$ 
\begin{equation}
\label{update_step1}
B_{t}(w_{i} \vert \theta_{t-1}) = \sum_{j} T(w_{j} \rightarrow w_{i}) B_{t-1}(w_{j} \vert \theta_{t-1})
\end{equation}
where $T(w_{j} \rightarrow w_{i})$ is the switching probability from subtask $j$ to subtask $i$.

Next, we compute the posterior belief distribution using Bayesian update, after observing the human action $a$ and subtask state $s_{i,t}$ at time $t$, as follows: 
\begin{equation}
\label{update_step2}
B_{t}(w_{i} \vert a_{t}=a,s_t, \theta_{t-1}) = \alpha . B_{t}(w_{i} \vert \theta_{t-1}) . P(a_{t} = a \vert w_{i}, s_{i,t})
\end{equation}
where  $\alpha$ is a normalizing constant. Absorbing current human action $a$ and current state into $\theta_{t-1}$ gives us the game history $\theta_{t}$ at time $t$.

\subsubsection{Complexity}

This component is run in real time, and thus its complexity dictates how responsive our AI is. We are going to show that it is at most $O(k^{2})$, with $k$ being the number of subtasks.

The first update step as depicted in Equation~\ref{update_step1} is executed for all subtasks, thus of complexity $O(k^{2})$. 

The second update step as of Equation~\ref{update_step2} requires the computation of $P(a_{t} = a \vert w_{i}, s_i)$ (Equation~\ref{update_actsel}), which takes $O(\vert A \vert)$ with $A$ being the set of compound actions. Since Equation~\ref{update_step2} is applied for all subtasks, that sums up to $O(k\vert A \vert)$ for this second step.

In total, the complexity of our real-time Intention Recognition component is $O(k^{2} + k\vert A \vert)$, which will be dominated by the first term $O(k^{2})$ if the action set is fixed. 

\subsection{Decision-theoretic Action Selection}
\label{sec:utility_sel}

Given a belief distribution on the player’s targeted subtasks as well as knowledge to act collaboratively optimally on each of the subtasks,  the agent chooses the action that maximizes its expected reward.
\[ a^{*} = \mbox{argmax}_{a} \left\lbrace \sum_{i} B_{t}(w_{i} \vert \theta_{t}) Q_{i}(s_{t}^{i}, a) \right\rbrace \]

\section{CAPIR: Collaborative Action Planner with Intention Recognition}
\label{sec:approach_imp}
We implement the scalable decision theoretic framework as a toolkit for implementing collaborative games, called Collaborative Action Planner with Intention Recognition (CAPIR).

\section{CAPIR's Architecture}

\begin{figure}[h]
\centering
\includegraphics[scale=0.5]{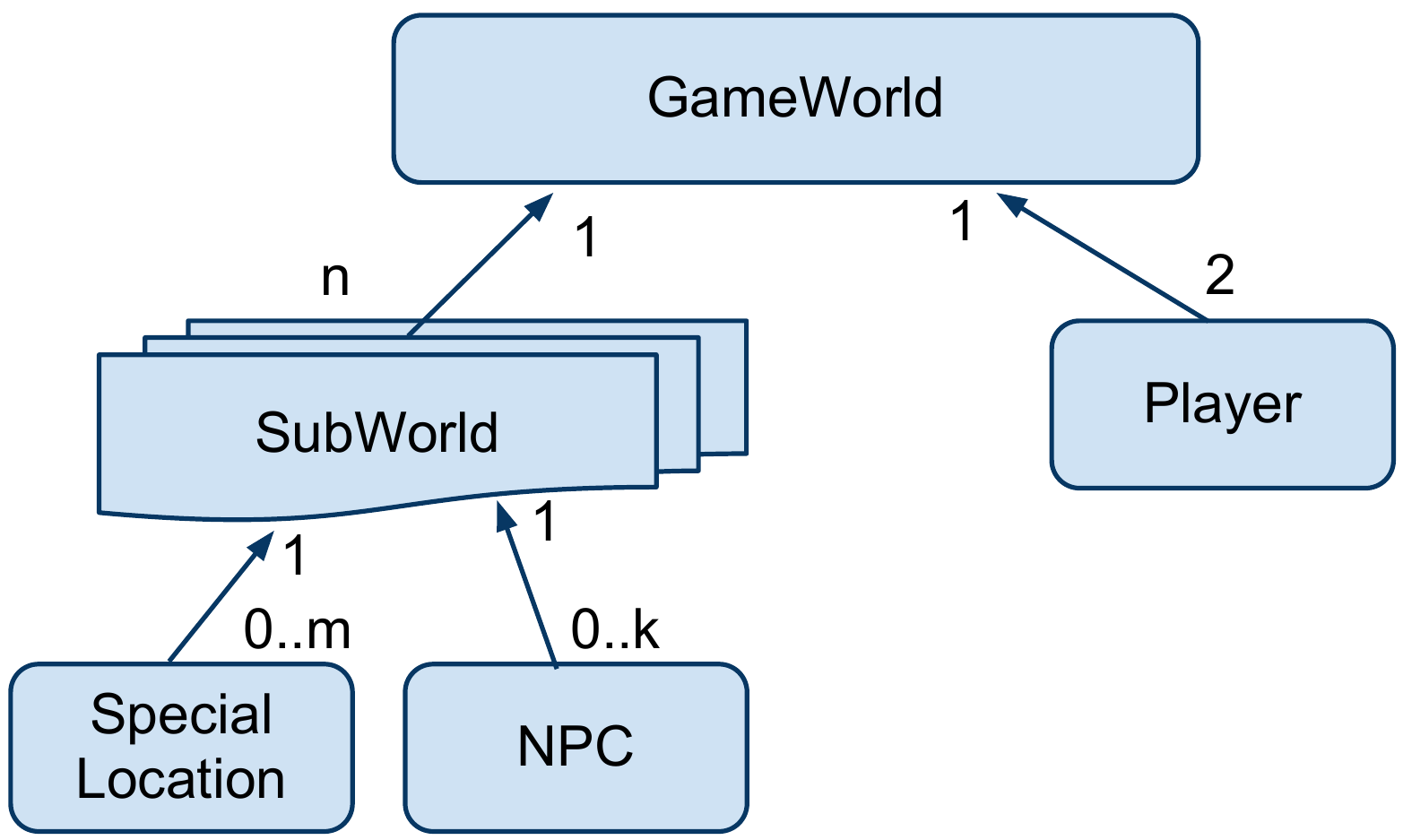}
\caption{GameWorld's components.}
\label{fig:capir_arch}
\end{figure}

Each game level in CAPIR is represented by a GameWorld object, which consists of two Players and multiple SubWorld objects, each of which contains only the elements required for a subtask (Figure~\ref{fig:capir_arch}). The game objective is typically to interact with these NPCs in such a way that gives the players the most points in the shortest given time. The players are given points in major events such as successfully killing a monster-type NPC or saving a civilian-type NPC -- these typically form the subtasks.

Each character in the game, be it the NPC or the protagonist, is defined in a class of its own, capable of executing multiple actions and possessing none or many properties. Besides movable NPCs, immobile items, such as doors or shovels, are specified by the class SpecialLocation. GameWorld maintains and updates an internal game state that captures the properties of all objects. 

\begin{figure}[h]
\centering
\includegraphics[scale=0.6]{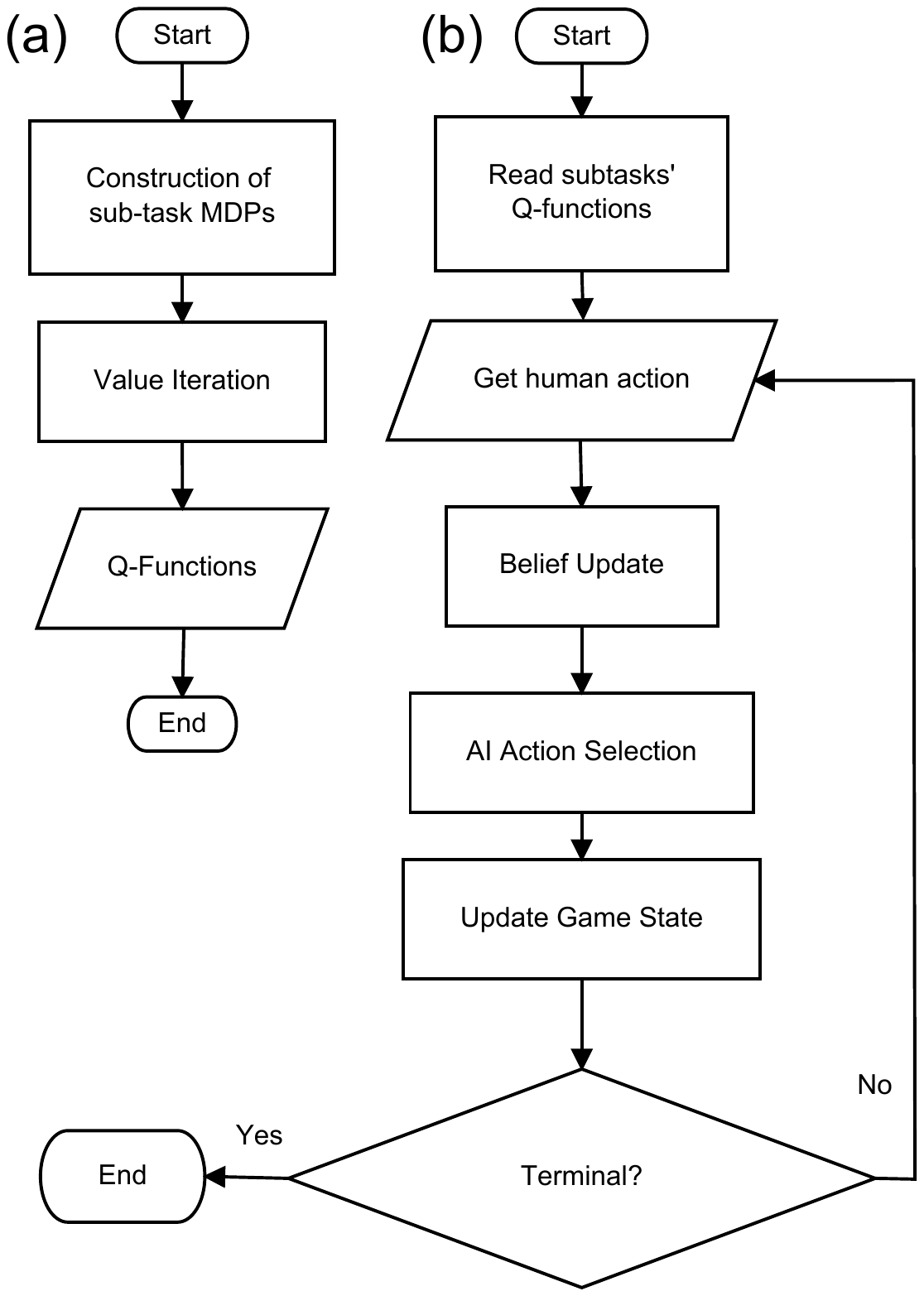} 
\caption{CAPIR's action planning process. (a) Offline subtask Planning, (b) in-game action selection using Intention Recognition.}
\label{fig:capir_pre}
\end{figure}

At the planning stage, for each SubWorld, an MDP is generated and a collaboratively optimal action policy is accordingly computed (Figure~\ref{fig:capir_pre}). These policies are used by the AI assistant at runtime to determine the most appropriate action to carry out, from a decision-theoretic viewpoint.

\section{Experiment and Analysis}

In order to evaluate the performance of our AI system, we conducted a human experiment using Collaborative Ghostbusters. We chose five levels (see Appendix) with roughly increasing state space size and game play complexity to assess how the technique can scale with respect to these dimensions. 

The participants were requested to play five levels of the game as Shepherd twice, each time with a helping Dog controlled by either AI or a member of our team, the so-called human expert in playing the game. The identity of the dog's controller was randomized and hidden from the participants. After each level, the participants were asked to compare the assistant's performance between two trials in terms of usefulness, without knowing who controlled the assistant at which turn.

In this set of experiments, the player's aim is to kill three ghosts in a maze, with the help of the assistant dog. The ghosts stochastically\footnote{The ghosts run away 90\% of the times and perform some random actions in the remaining 10\%.} run away from any protagonists if they are $4$ steps away. At any point of time, the protagonists could move to an adjacent free grid square or shoot; however, the ghosts only take damage from the ghost-buster if he is $3$ steps away. This condition forces the players to collaborate in order to win the game. In fact, when we try the game with non-collaborative dog models such as random movement, the result purely relies on chance and could go on until the time limit (300 steps) runs out, as the human player hopelessly chases ghosts around obstacles while the dog is doing some nonsense at a corner. Oftentimes the game ends when ghosts walk themselves into dead-end corners. 

The twenty participants are all graduate students at our school, seven of whom rarely play games, ten once to twice a week, and three more often.

When we match the answers back to respective controllers, the comparison results take on one of three possible values, being AI assistant performing ``better", ``worse" or ``indistinguishable" to the human counterpart. The AI assistant is given a score of 1 for a ``better", 0 for an ``indistinguishable" and -1 for a ``worse" evaluation. 

\subsubsection{Qualitative evaluation}
For simpler levels 1, 2 and 3, our AI was rated to be better or equally good more than 50\% the times. For level 4, our AI rarely got the rating of being indistinguishable, though still managed to get a fairly competitive performance. Subsequently, we realized that in this particular level, the map layout is confusing for the dog to infer the human's intention; there is a trajectory along which the human player's movement could appear to aim at any one of three ghosts. In that case, the dog's initial subtask belief plays a crucial role in determining which ghost it thinks the human is targeting. Since the dog's belief is always initialized to a uniform distribution, that causes the confusion. If the human player decides to move on a different path, the AI dog is able to efficiently assist him, thus getting good ratings instead. In level 5, our AI gets good ratings only for less than one third of the times, but if we count ``indistinguishable" ratings as satisfactory, the overall percentage of positive ratings exceeds 50\%.

\begin{figure}[h]
\centering
\includegraphics[scale=0.60]{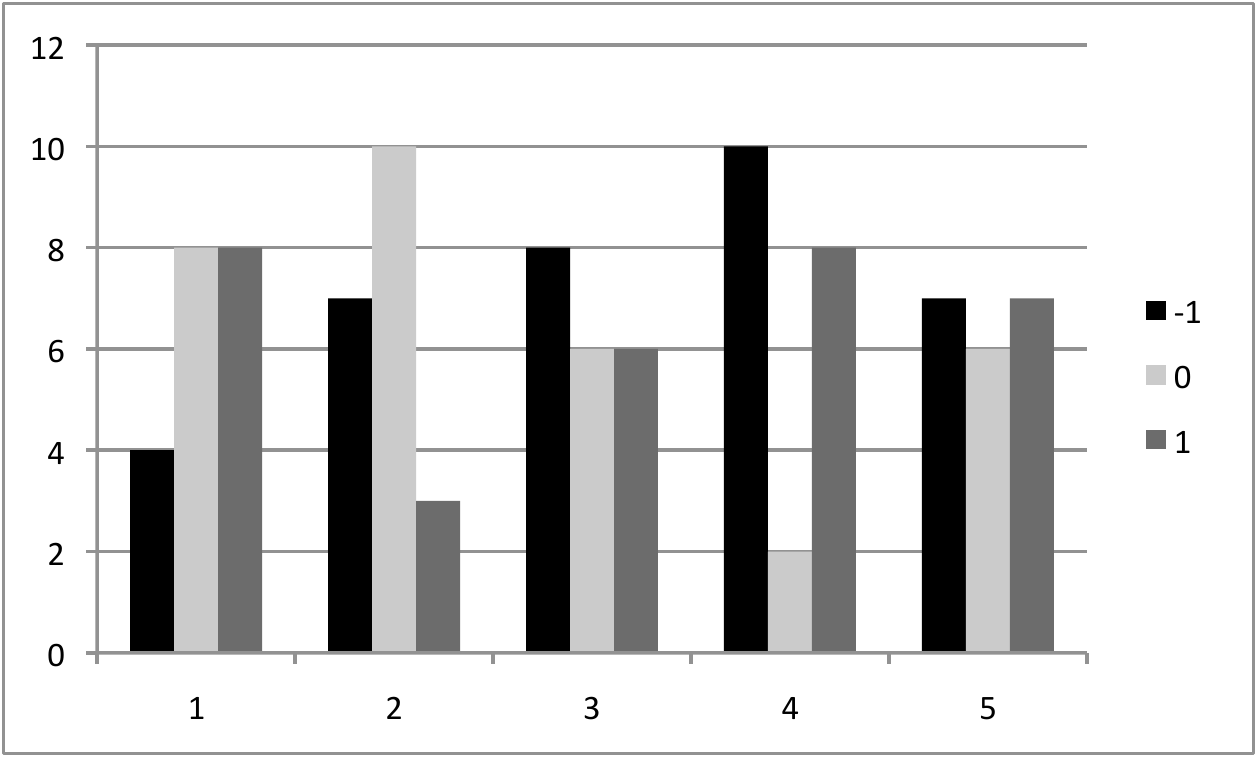}
\caption{Qualitative comparison between CAPIR's AI assistant and human expert. The y-axis denotes the number of ratings.}
\label{fig:human_exp}
\end{figure}

\begin{figure}[h]
\centering
\includegraphics[scale=0.64]{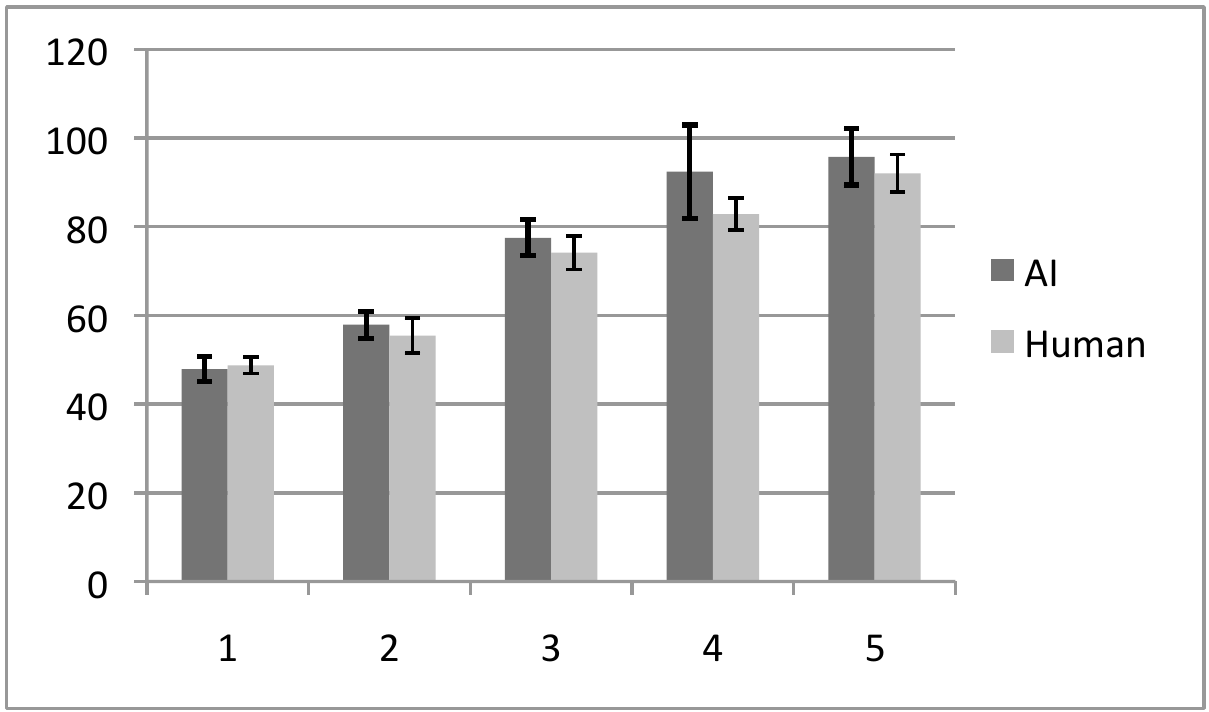}
\caption{Average time, with standard error of the mean as error bars, taken to finish each level when the partner is AI or human. The y-axis denotes the number of game turns.}
\label{fig:3rd_ghost}
\end{figure}

\subsubsection{Quantitative evaluation}

Besides qualitative evaluation, we also recorded the time taken for participants to finish each level (Figure~\ref{fig:3rd_ghost}). Intuitively, a well-cooperative pair of players should be able to complete Collaborative Ghostbuster's levels in shorter time. 

Similar to our qualitative result, in levels 1, 2 and 3, the AI controlled dog is able to perform at near-human levels in terms of game completion time. Level 4, which takes the AI dog and human player more time on average and with higher fluctuation, is known to cause confusion to the AI assistant's initial inference of the human's intention and it takes a number of game turns before the AI realizes the true target, whereas our human expert is quicker in closing down on the intended ghost. Level 5, larger and with more escape points for the ghosts but less ambiguous, takes the protagonist pair (AI, human) only 4.3\% more on average completion time. 

\section{Related Work}

Since plan recognition was identified as a problem on its own right in 1978~\cite{Schmidt1978}, there have been various efforts to solve its variant in different domains. In the context of modern game AI research, Bayesian-based plan recognition has been inspected using different techniques such as Input Output Hidden Markov Models~\cite{gold2010}, Plan Networks~\cite{orkin2007restaurant}, text pattern-matching~\cite{mateas2007writing}, n-gram and Bayesian networks~\cite{mott2006probabilistic} and dynamic Bayesian networks~\cite{albrecht1998}. As far as we know, our work is the first to use a combination of precomputed MDP action policies and online Bayesian belief update to solve the same problem in a collaborative game setting.

Related to our work in the collaborative setting is the work reported by Fern and Tadepalli ~\cite{fern2010computational} who proposed a decision-theoretic framework of assistance. There are however several fundamental differences between their targeted problem and ours. Firstly, they assume the task can be finished by the main subject without any help from the AI assistant. This is not the case in our game, which presents many scenarios in which the effort from one lone player would amount to nothing and a good collaboration is necessary to close down on the enemies. Secondly, they assume a stationary human intention model, i.e. the human only has one goal in mind from the start to the end of one episode, and it is the assistant's task to identify this sole intention. In contrary, our engine allows for a more dynamic human intention model and does not impose a restriction on the freedom of the human player to change his mind mid way through the game. This helps ensure our AI's robustness when inferring the human partner's intention.

In a separate effort that also uses MDP as the game AI backbone, Tan and Cheng~\cite{chek2010} model the game experience as an abstracted MDP - POMDP couple. The MDP models the game world's dynamics; its solution establishes the optimal action policy that is used as the AI agent's base behaviors. The POMDP models the human play style; its solution provides the best abstract action policy given the human play style. The actions resulting from the two components are then merged; reinforcement learning is applied to choose an integrated action that has performed best thus far. This approach attempts to adapt to different human play styles to improve the AI agent's performance. In contrast, our work introduces the multi-subtask model with intention recognition to directly tackle the intractability issue of the game world's dynamics.

\section{Conclusions}
We describe a scalable decision theoretic approach for constructing collaborative games, using MDPs as subtasks and intention recognition to infer the subtask that the player is targeting at any time. Experiments show that the method is effective, giving near human-level performance.

In the future, we also plan to evaluate the system in more familiar commercial settings, using state-of-the-art game platforms such as UDK or Unity. These full-fledged systems offer development of more realistic games but at the same time introduce game environments that are much more complex to plan. While experimenting with Collaborative Ghostbuster, we have observed that even though Value Iteration is a simple naive approach, in most cases, it suffices, converging in reasonable time. The more serious issue is the state space size, as tabular representation of the states, reward and transition matrices takes much longer to construct. We plan to tackle this limitation in future by using function approximators in place of tabular representation.

\section{Acknowledgments}

This work was supported in part by MDA GAMBIT grant R-252-000-398-490 and AcRF grant T1-251RES0920 in Singapore. The authors would like to thank Qiao Li (NUS), Shari Haynes and Shawn Conrad (MIT) for their valuable feedbacks in improving the CAPIR engine, and the reviewers for their constructive criticism on the paper.

\bibliographystyle{aaai}
\bibliography{group-3876}


\renewcommand{\topfraction}{.95}
\renewcommand{\bottomfraction}{.95}
\renewcommand{\textfraction}{.05}
   
\appendix
\section{Appendix}
Game levels used for our experiments.

\begin{figure}[h]
  \centering
  \subfloat{\label{fig:level1}\includegraphics[scale=0.54]{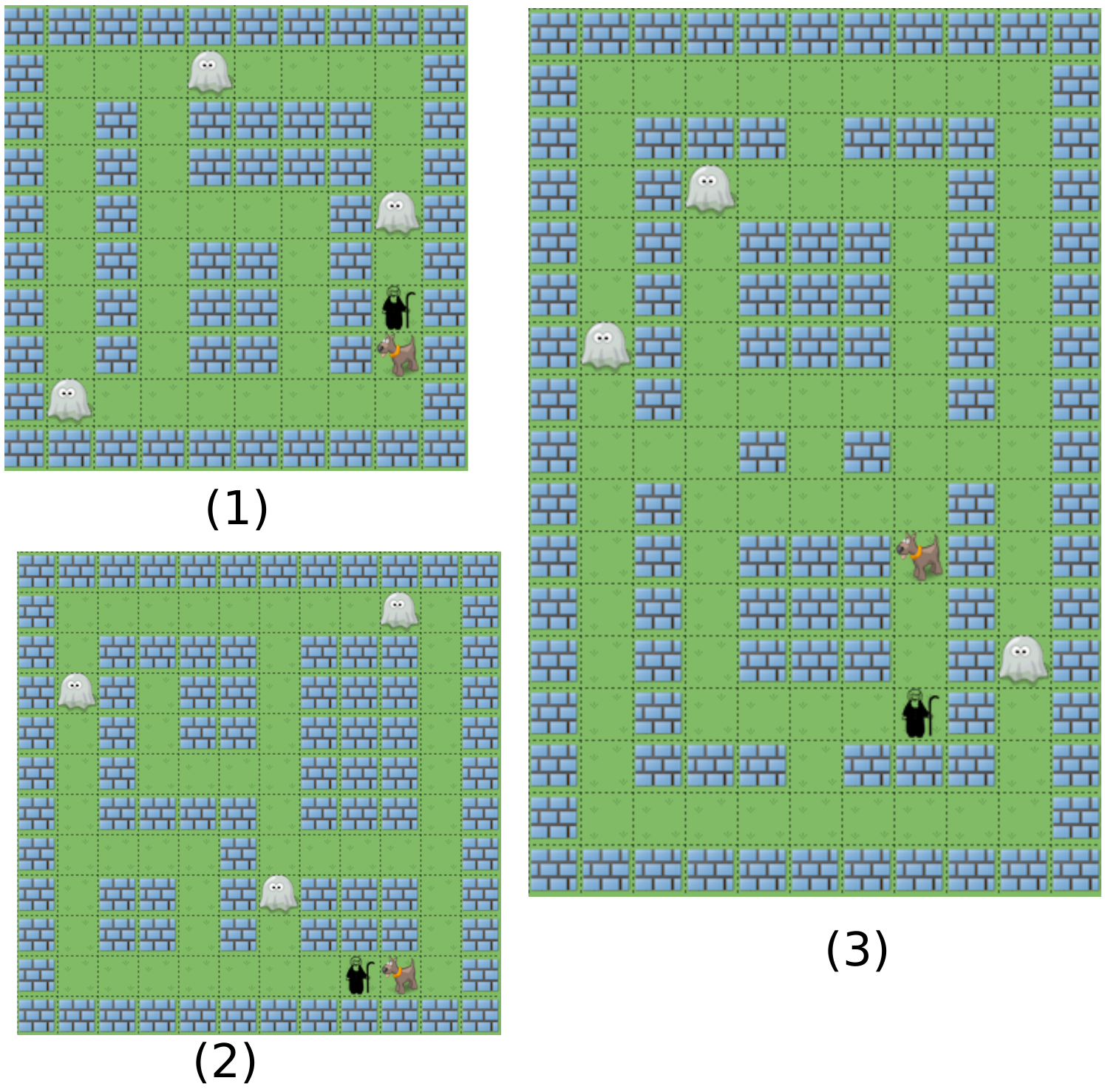}} 
  \\
  \subfloat{\label{fig:level2}\includegraphics[scale=0.6]{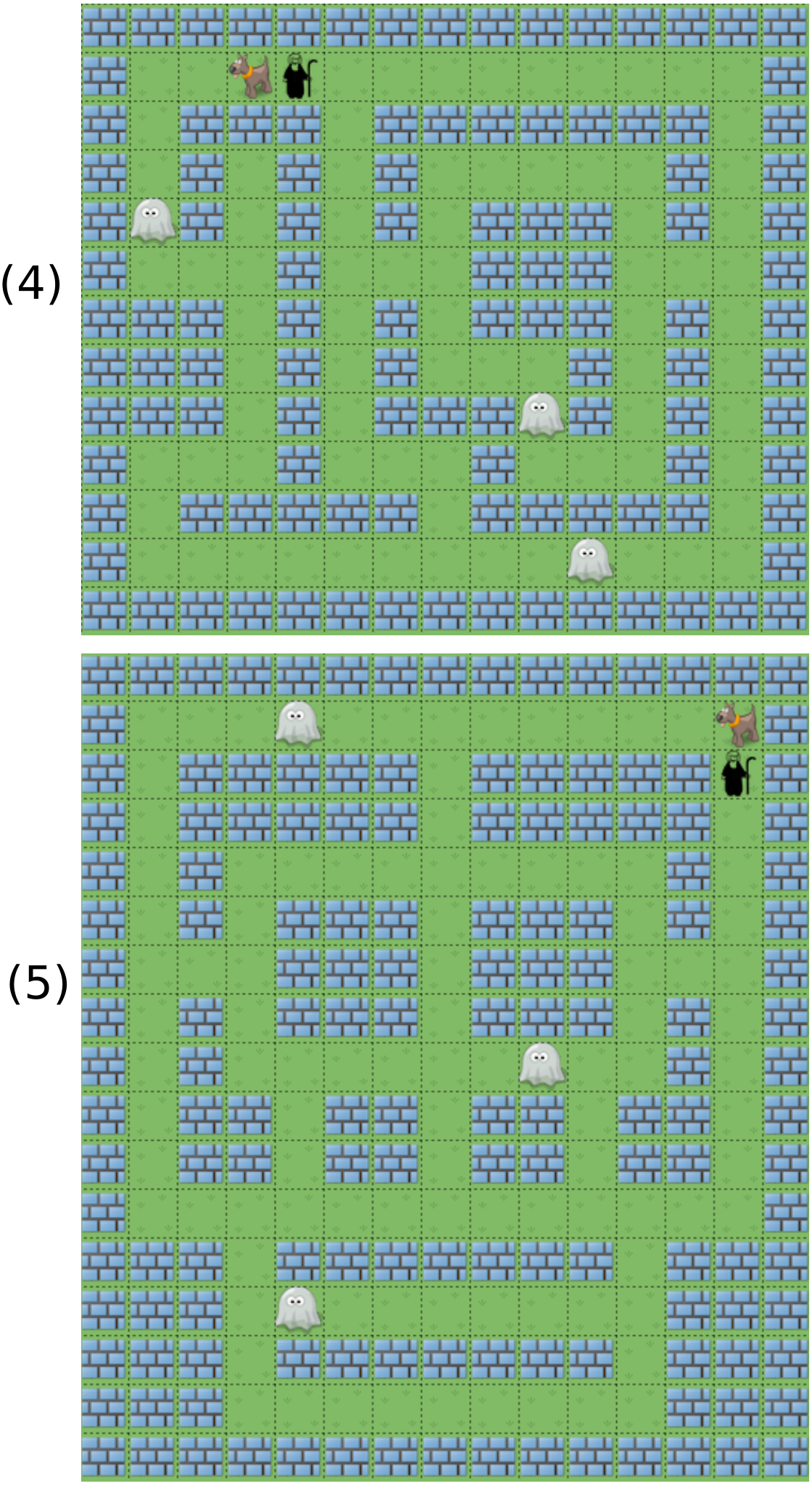}}
  \label{fig:animals}
\end{figure}
%
%

\end{document}